\documentclass[A4paper,11pt]{marine_2023_Paper}
\usepackage{url}
\usepackage{hyperref}
\usepackage{graphicx}
\usepackage{amsmath}
\usepackage{amsfonts}
\usepackage{amssymb}
\usepackage{titlesec}
\usepackage{mathtools,xparse}
\usepackage{todonotes}
\usepackage{regexpatch}
\titleformat{\section}[block]
{\normalfont\bfseries\filcenter}
{\thesection.}{.5em}{\bfseries}

\DeclarePairedDelimiter{\norm}{\lVert}{\rVert}
\usepackage{float}
\usepackage{enumitem}
\usepackage{subcaption}
\usepackage[backend=biber, sorting=nyt, bibstyle=apa, citestyle=authoryear, maxcitenames=1]{biblatex} 
\usepackage{setspace}
\DeclareDelimFormat[parencite]{nameyeardelim}{\addcomma\space}
\DeclareDelimFormat[textcite]{nameyeardelim}{%
  \ifnumgreater{\value{liststop}}{2}{\finalandcomma}{}\space}

\DefineBibliographyStrings{english}{
  andothers = {\mkbibemph{et\addabbrvspace al\adddot}}
}

\setuptodonotes{inline}

\title{Multi-Objective Hull Form Optimization with CAD Engine-based Deep Learning Physics  for 3D Flow Prediction}

\author{Jocelyn Ahmed Mazari$^{1*}$, Antoine Reverberi$^{1*}$, Pierre Yser$^{1*}$ and Sebastian Sigmund$^{\dag*}$}

\heading{Jocelyn Ahmed Mazari, Antoine Reverberi, Pierre Yser and Sebastian Sigmund}

\address{$^{1}$ Extrality, 75002, Paris, France, web page: www.extrality.ai
\and
$^{\dag}$ Bremen University of Applied Sciences, 28199, Bremen, Germany, web page: www.hs-bremen.de/en/person/sesigmund\\[.5cm]
$^{*}$ Corresponding authors: Jocelyn Ahmed Mazari, Antoine Reverberi, Pierre Yser and Sebastian Sigmund, emails: \{ahmed,antoine,pierre@extrality.ai\}, sebastian.sigmund@hs-bremen.de}

\begin{document}
\maketitle
\thispagestyle{empty}
\setlength{\parskip}{0.3cm}

\begin{center}{\bf ABSTRACT}\end{center}
In this work, we propose a built-in \textit{Deep Learning Physics Optimization (DLPO)} framework to set up a shape optimization study of the Duisburg Test Case (DTC) container vessel. We 
present two different applications: (1) sensitivity analysis to detect the most promising generic basis hull shapes, and (2) multi-objective optimization to quantify the trade-off between optimal hull forms. DLPO framework allows for the evaluation of design iterations automatically in an end-to-end manner. 
We achieved these results by coupling Extrality's \textit{Deep Learning Physics (DLP)} model to a CAD engine and an optimizer. 
Our proposed DLP model is trained on full 3D volume data coming from RANS simulations, and it can provide accurate and high-quality 3D flow predictions in real-time, which makes it a good evaluator to perform optimization of new container vessel designs w.r.t the hydrodynamic efficiency. In particular, it is able to recover the forces acting on the vessel by integration on the hull surface with a mean relative error of $3.84\% \pm 2.179\%$ on the total resistance. Each iteration takes only 20 seconds, thus leading to a drastic saving of time and engineering efforts, while delivering valuable insight into the performance of the vessel, including RANS-like detailed flow information. We conclude that DLPO framework is a promising 
tool to accelerate the ship design process and lead to more efficient ships with better hydrodynamic performance.


\section*{NOMENCLATURE}
\begin{tabular}{ll}
ANNs & Artificial Neural Networks \\
CAD & Computer-Aided Design \\
CFD & Computational Fluid Dynamics \\
DACE  & Design and Analysis of Computer Experiment \\
DL & Deep Learning \\
DLP & Deep Learning Physics \\
DLPO & Deep Learning Physics Optimization\\
DTC & Duisburg Test Case \\
HSB & Bremen University of Applied Sciences \\
GDL & Geometric Deep Learning \\
MAE & Mean Absolute Error \\
ML & Machine Learning \\
NURBS &  Non-Uniform Rational B-Spline \\
RANS & Reynolds-Averaged Navier-Stokes \\
SGD & Stochastic Gradient Descent \\
T-Search & Tangent Search \\
\end{tabular}


\section{INTRODUCTION}\label{sec:introduction}

\subsection{Related works}\label{subsec:state_of_the_art}
\noindent {\bf Optimization methods}. Existing optimization methods based on Computational Fluid Dynamics (CFD) solutions require the evaluation of a large number of designs which could be expensive when computed with numerical solvers and is intractable in practice. To reduce the number of required simulations, a well-known approach is to construct a \textit{meta-model} or \textit{surrogate} model \parencite{forrester2018}, which is based on a set of training simulations and can provide an approximation of the simulation solution for any parameter configuration in a reasonable time. Therefore, it enables engineers to iterate on the model as many times as needed to optimize their designs.
In \textcite{queipo2005}, a comparative study on surrogate-based optimizations is introduced including several approaches namely the design of experiments, surrogate selection and construction, sensitivity analysis, and convergence, as well as optimization.
Surrogate modeling techniques have been widely used in ship design and optimization. \textcite{scholcz2015} use surrogates to obtain an approximation of the \textit{Pareto fronts} from the hull form optimization of a ship, including free-surface effects. It is shown that the ship design process can be accelerated, leading to more efficient ships. In \textcite{scholcz2017}, the authors demonstrate that surrogate techniques are an efficient way
to mitigate the computational cost during the optimization phase. For a practical application, they 
show that the computational time can be reduced from \textit{two weeks to one day} when using a surrogate-based global optimization technique instead of applying a multi-objective optimization directly to the solver.

\noindent {\bf Artificial Neural Networks (ANNs)}.
ANNs have been used in marine applications for three decades \parencite{sha1994}. They are applied to tackle various aspects of the ship design process thanks to their universal approximation properties that enable them to approximate a wide range of continuous functions \parencite{hornik1989}. Since 2012, the resurgence of ANNs as Deep Learning (DL) \parencite{krizhevsky2012}
has opened new perspectives for CFD thanks to the availablability of a huge quantity of data \parencite{pfaff2020}. For design optimization, DL can also be used
to build surrogate models as an alternative to commonly used approaches such as quadratic and cubic polynomial surfaces, or kriging \parencite{raven2017}.
In \textcite{yu2018}, DL is used to accurately establish the relationship between different hull forms and their associated performances. The reported results show that large quantities of different hull forms can be evaluated at negligible computing costs and with a similar level of accuracy as the numerical solver itself. Therefore, a DL-based optimization can be performed with 
efficient speed and flexibility. 
Another study conducted in \textcite{vandenboogaard2022} proposes a DL model to predict both integral quantities of open-water performance curves and local field values of velocities and pressures on two-dimensional planes for a propeller family. The results show that the proposed DL model is capable of predicting the local flow fields and its generalizability, as well as its quasi-instantaneous response may accelerate propeller design significantly.

\noindent {\bf Overlapping design and simulation activities}. With the development of CFD, the prediction and analysis of the hydrodynamic efficiency of a ship have been more and more the responsibility 
of CFD experts. In \textcite{vanstraten2019}, the study emphasizes the importance of transferring at least a part of this analysis and computation-based design trade-off back to the ship designer. They showed the role that the Design and Analysis of Computer Experiments (DACE) methodology can play in the early design phase of a ship and how to, relatively quickly, set up a surrogate model to gather valuable insight into the performance of the ship, including multi-objective shape optimization and inverse design.

Based on the literature review, several conclusions can be made:
\begin{itemize}
    \item Optimization methods based on CFD often rely on the construction of a surrogate model 
    to get an accurate prediction of the CFD output 
    within a short amount of time and without requiring a new, expensive evaluation. 
    This technique allows the evaluation directly on the shape parameters instead of the hull shape itself. This is not enough to provide a suitable representation of the complex input/output relation and it cannot easily be mapped to an entirely different ship type.
    \item In previous works, ANNs have been extensively used for ship design and optimization. Despite this interest, there is a very limited number of geometry-centric applications. Additionally, these methods fail to exploit the 3D flow field generated by the CFD, thus providing insufficient information about the performance of the ship.
    \item Currently, there is a need to increase the overlap between the simulation expert and the ship designer. A Computer-Aided Design (CAD) engine-based 3D surrogate model would enable the ship designer to (i) rapidly explore the consequences of design choices, (ii) reduce the through-put time of a project, and (iii) reduce the workload on the simulation 
    specialist.
\end{itemize}

\subsection{Our contribution}\label{subsec:our_contribution}

In this work, we propose to build a physics-based DL model that we denote \textit{Deep Learning Physics (DLP)} as a surrogate model for ship design.
Our DLP is able to learn from the full 3D volume data coming from RANS (Reynolds-Averaged Navier-Stokes) CFD simulations and make predictions at different scales, including total forces (global coefficients),
distribution of forces on the hull (surface fields), 3D flow field values (volume fields). The goal of our DLP is to enhance the hydrodynamic efficiency around the vessels and ultimately can be used to optimize new container vessel designs that will be described in the next section \ref{subsec:shape_optim}. Moreover, our DLP model works on a topological representation of the geometry, which allows us to extract meaningful features and to accurately compute the force coefficients acting over geometries. This provides an increased understanding of the input/output relation, in particular in the early design phase of a ship.

Recent efforts have successfully given rise to the development of efficient 
engineering design approaches, allowing the designers to rapidly explore the performance of a large number of hull variations while ensuring high accuracy and quality of the predictions. Such developments have created more overlap between the simulation expert and the ship designer in the knowledge-value chain. By coupling our DLP model to a 
CAD engine and an optimizer, we propose a built-in \textit{Deep Learning Physics Optimization (DLPO)} framework to achieve efficient shape optimization.

\subsection{Structure of the paper}\label{subsec:structure_of_the_paper}

The paper is structured as follows: in section \ref{sec:methodology} we present our DLP and optimization framework that allow making automatic shape deformation and optimization. Then, in section \ref{sec:experiments} we describe the experimental protocol and evaluate the performances of the proposed framework. Finally, in section \ref{sec:conclusions} we conclude our study with a summary and some open directions.

    


\section{METHODOLOGY}\label{sec:methodology}

In this section, we present our DLPO framework. First, a general description of the DLP model training is provided. Then, we show how to build a multi-objective optimization framework based on the outputs of the DLP model.

\subsection{Deep Learning Physics (DLP) architecture}\label{subsec:dlp_framework}


\noindent {\bf Mathematical model}. Machine Learning (ML) is a high-dimensional interpolation problem. 
In this study, our ML task is defined as \textit{multivariate regression} problem where our DLP is supervised with the solution of the CFD solver.
We approximate a function $f: X \in \mathbb{R}^{3} \rightarrow Y \in \mathbb{R}^{m}$ from a collection of $n$ flow simulations $\mathcal{D}=\{(g_{i},(V_{\infty})_{i},\mathcal{A}_{i})\}_{i \leq n}$. $x \longmapsto f(x)$ maps a point in $X$ to a set of $m$ variables. $X$ is the set of all points in the volume of simulation and $f(x)=(\overline{p}(x),\overline{q}(x),\overline{u}(x), \overline{\kappa}(x))$
the spatial fields to be predicted; namely the pressure $\overline{p}$, water volume fraction $\overline{q}$, the mean-field velocity  $\overline{u}=\{\overline{u}_{x},\overline{u}_{y},\overline{u}_{z}\}$, and the skin friction components $\overline{\kappa}=\{\overline{\kappa}_{x},\overline{\kappa}_{y},\overline{\kappa}_{z}\}$. $g_{i}$ represents the geometry and is defined as 3D mesh, $(V_{\infty})_{i}$ is the freestream condition for $g_{i}$ and $\mathcal{A}_{i}=\{(x_{k},f_{i}(x_{k})\}_{k=1}^{K_{i}}$ a cloud of points induced by $g_{i}$ and $(V_{\infty})_{i}$.

We denote $\mathcal{F}$ as the space of functions that maps $X$ to $Y$. Hence, the learning problem here consists of finding the function $\hat{h} \in \mathcal{F}$ that best fits the dataset of simulations $\mathcal{D}$ and generalizes well to new geometries $g$ and new freestream conditions $V_{\infty}$. The optimal model $\hat{h}$ is expressed as follows:

\begin{equation}
    \hat{h}= \underset{h \in \mathcal{F}}{\mathrm{argmin}} \ \mathcal{L}(h,\mathcal{D})
\end{equation}

With $\mathcal{L}$ a regularized loss function defined over $\mathcal{D}$ that measures the errors of the model. In other terms, it evaluates how well the learning algorithm is modeling the dataset $\mathcal{D}$. In this work, we use $L1$ loss that we denote $\ell(.)$ known as \textit{Mean Absolute Error} (MAE) which is the absolute difference between the predictions of the DLP model and the ground truth (solution of CFD solver).

Again, our DLP model aims at learning a function that takes as input a geometry $g$, a freestream condition $V_{\infty}$ to regress the aforementioned variables defined at $x \in X \ \forall x$. This function is approximated by a neural network that we denote $h_{\theta_{pde}}$ with $\theta_{pde}$ a set of trainable parameters to be optimized during the training time using \textit{Stochastic Gradient Descent} (SGD) \parencite{kingma2014} as follows:

\begin{equation}
    \begin{split}
        \hat{\theta} & = \underset{\theta}{\mathrm{argmin}} \sum_{i=1}^{n} \sum_{k=1}^{K_{i}} \ell(f_{i}(x_{k}),h_{\theta_{pde}}(x_{k}|g_{i},(V_{\infty})_{i})) \\
        & = \underset{\theta}{\mathrm{argmin}} \left[ \underbrace{\sum_{i=1}^{n} \sum_{k=1}^{K_{i} \in  \mathcal{V}}  \norm{\overline{p_{v}^{i}}(x_{k}) - \overline{\widehat{p}_{v}^{i}}(x_{k})}_{1} + \norm{\overline{u^{i}}(x_{k}) - \overline{\widehat{u}^{i}}(x_{k})}_{1} + \norm{\overline{q_{v}^{i}}(x_{k}) - \overline{\widehat{q}_{v}^{i}}(x_{k})}_{1}}_{\text{loss over the volume \ $x \in \mathcal{V}$}} \right] + \\
        & \underset{\theta}{\mathrm{argmin}} \left[ \underbrace{\sum_{i=1}^{n} \sum_{k=1}^{K_{i} \in  \mathcal{S}}  \norm{\overline{p_{s}^{i}}(x_{k}) - \overline{\widehat{p}_{s}^{i}}(x_{k})}_{1}  + \norm{\overline{q_{s}^{i}}(x_{k}) - \overline{\widehat{q}_{s}^{i}}(x_{k})}_{1} + \norm{\overline{\kappa^{i}}(x_{k}) - \overline{\widehat{\kappa}^{i}}(x_{k})}_{1}}_{\text{Loss over the surface  \ $x \in \mathcal{S}$}} \right]
    \end{split}
\end{equation}

Our DLP model regresses the following spatial physical fields (CFD solution ground truths) $\overline{p}=\{\overline{p}_{v},\overline{p}_{s}\}$, $\overline{u}=\{\overline{u}_{x},\overline{u}_{y},\overline{u}_{z}\}$, $\overline{q}=\{\overline{q}_{v},\overline{q}_{s}\}$, and $\overline{\kappa}=\{\overline{\kappa}_{x},\overline{\kappa}_{y},\overline{\kappa}_{z}\}$ with $v \in \mathcal{V} \subset{X}, \ s \in \mathcal{D} \subset{X}$ stand respectively for volume and surface points, and $\overline{\widehat{p}_{v}}, \overline{\widehat{u}}, \overline{\widehat{q}_{v}}, \overline{\widehat{p}_{s}}, \overline{\widehat{q}_{s}}, \overline{\widehat{\kappa}}$  the predictions of the model $h_{\theta_{pde}}$. Then, the 
total resistance is computed as post-treatment (derived quantity from $h_{\theta_{pde}}$).

The architecture of our DLP model is derived from the Geometric Deep Learning (GDL) framework \parencite{bronstein2017} and is endowed with physics knowledge to achieve learning on the full 3D volume data (see later in section \ref{subsec:dlp_exp}). It can learn on arbitrary (irregular) meshes with complex geometrical variations while being robust to the sampling. It could be trained on coarse meshes while providing high-quality results on (new) fine-grained meshes unseen during the training.





\subsection{Deep Learning Physics Optimization (DLPO) framework}\label{subsec:shape_optim}

\noindent {\bf Practical applications}. After constructing the DLP model, it can be used for various optimization purposes, such as design exploration and design exploitation. During the exploration phase, a number of shape variants is created and analyzed with the goal of understanding the influence and the relation between the parameters variations and their impact on the performances. However, the number of individual simulations that are required to obtain statistically significant trends can increase exponentially w.r.t. the design directions to be looked at.
Such effect is known as the \textit{curse of dimensionality} \parencite{forrester2018}. Therefore, a sensitivity study is conducted to detect the most promising generic basis hull shapes, particularly to distinguish parameters of higher importance from those with less influence.
As a second step, an exploitation phase follows to squeeze out the best possible results. It is commonly done within reduced regions, to look for local optima, or within a subspace in which some of the less important free variables are frozen. Finally, selected variants are analyzed and compared to the baseline, possibly at higher grid resolution and for conditions (and in disciplines) not considered during the optimization. 
From those variants the most favorite one is selected, 
concluding the 
design study.

\begin{figure}[h]
    \includegraphics[width=\linewidth]{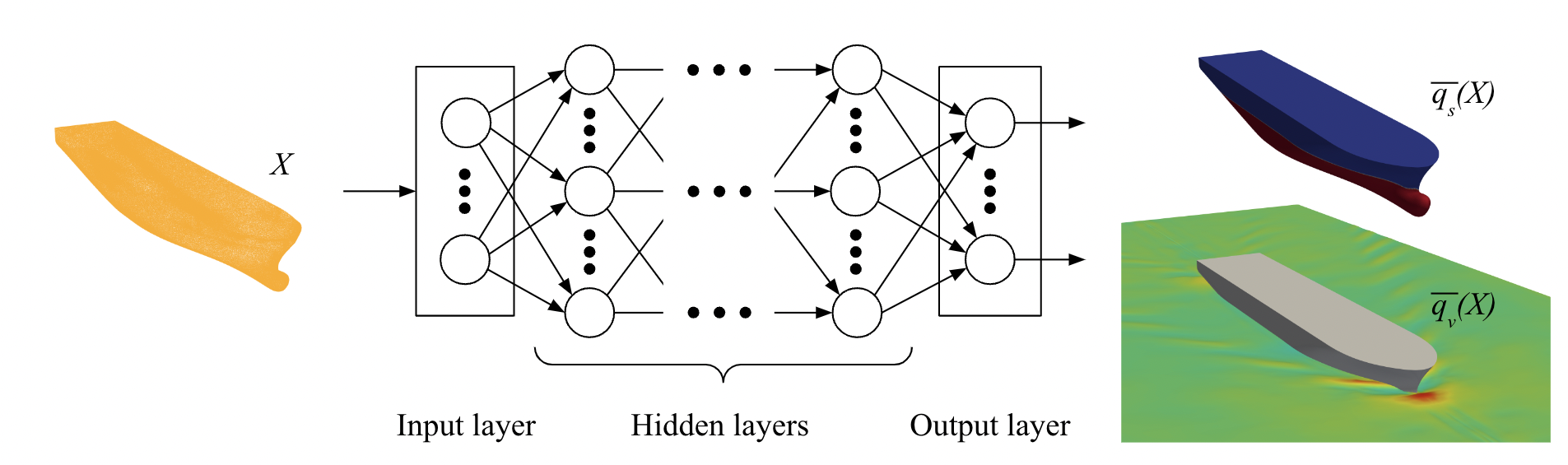}%
    \caption{The architecture of our Deep Learning Physics (DLP) model. 
    Once the DLP model is trained, it is able to associate a RANS-like 3D flow prediction (right) to a new, unseen 3D mesh and freestream condition (left). As an example of output, we show the surface and volume field of water volume fraction (respectively $\overline{q_{s}}$ and $\overline{q_{v}}$).}
    \label{fig:ann}
\end{figure}

In this work, we present two different applications, as an example, to show the capabilities of DLP model in shape optimization: (1) \textit{sensitivity analysis} and (2) \textit{multi-objective optimization}. First, in the sensitivity analysis study, we expose the input/output relation by screening the design space and ranking the design parameters. Secondly, in multi-objective optimization study, we visualize and/or quantify the trade-off between designing optimal hull forms for different objectives. The output is the surface and volume variable predictions, as well as the 
total forces acting on the vessel.

\noindent {\bf Hull shape representation}. In shape deformation and optimization studies, a method to automatically deform the surface or mesh representation of a shape is required. In CAD software, the standard description used to describe shapes are Non-Uniform Rational B-Spline (NURBS) curves and surfaces \parencite{letcher2009}.
A NURBS curve is defined by a set of weighted control points, and a knot vector, as well as the order of the curve.
Effectively, a NURBS surface is the tensor product of two NURBS curves. Deforming a NURBS curve and surface to automatically generate new shapes can be achieved by changing either the control points, knot vector or order. However, it is challenging due to the large number of degrees of freedom. For CFD computation, the object geometry is represented by a 
mesh. 
Mesh deformation is done by adjusting and parametrizing the mesh of a simulation model. However, it is difficult to recover the CAD model from the optimized mesh as it involves that each surface 
needs to be suitably represented. 
As a consequence, the hull should undergo a fairing process\footnote{Hull fairing is a major point of hull design. A faired hull can be described as smooth and free from bumps or discontinuities \parencite{britannica2018}.} before it can be 
used effectively for further design work. 
Another solution consists in dealing with the parametric deformation of an existing, imported geometry directly within the CAD software. This method, called \textit{geometric morphing}, allows to generate shapes easily while processing the shapes for maximum fairness. Thus, all shapes are created based on given parameters and they are generated instantly when a parameter is modified.

\begin{figure}[h]
    \includegraphics[width=\linewidth]{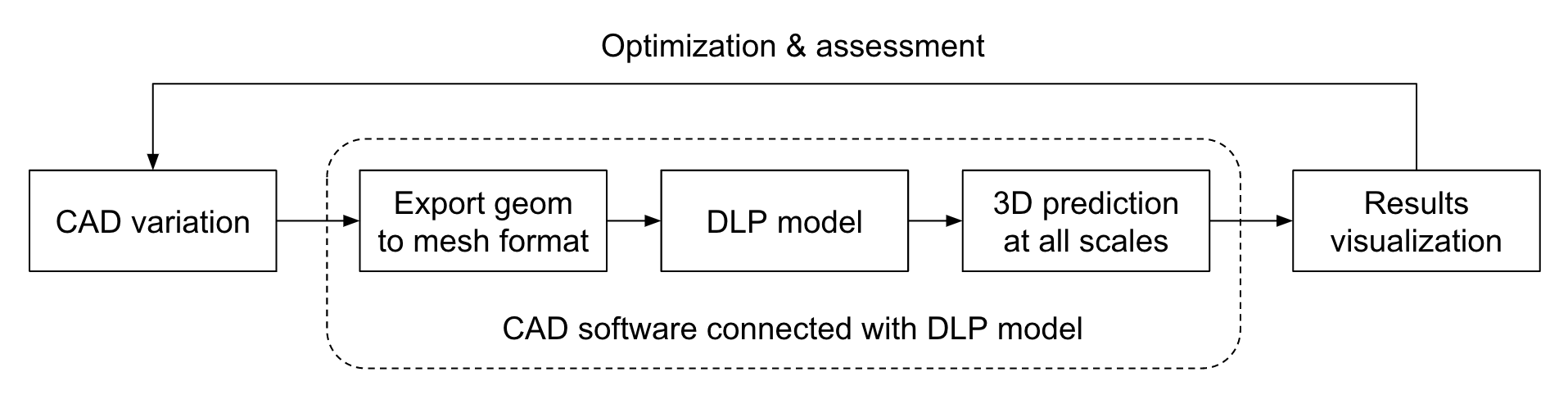}%
    \caption{The proposed Deep Learning Physics Optimization (DLPO) framework is the result of linking the DLP model with a CAD software in an optimization loop. The DLP model provides 3D predictions at all scales: surface and volume fields, as well as total forces.}\label{fig:optim-process}
\end{figure}

In order to set up the optimization framework, it is necessary to couple the CAD modeler, which drives the selection of geometry variants, to the DLP model, which provides the predictive performance results. When properly coupled, the CAD modeler runs a DLP prediction to evaluate design iterations automatically and the results of the analyses are 
subsequently 
used to set up a fast optimization algorithm. The whole optimization pipeline is illustrated in Figure \ref{fig:optim-process}.


In this work, we use the Sobol sequence for exploration and the Tangent Search (T-Search) method for exploitation. The Sobol sequence is a statistical method to generate a quasi-random range of samples from a multi-dimensional distribution \parencite{sobol1988}. It tries to fill up the design space as uniformly as possible with as few variants as possible. The T-Search method is a gradient-free method combining smaller steps and larger moves through the design space that can also handle inequality constraints \parencite{hilleary1966}. See later on, the applications of theses methods in section \ref{subsec:exp_shape_optim}.


\section{EXPERIMENTS}\label{sec:experiments}


In this section, we present the results of the DLPO framework. To do so, we first describe the dataset used to build the DLP model. Secondly, we report quantitative and qualitative results to assess the performances of the DLP model. Finally, we show two applications of the DLPO framework: (1) \textit{sensitivity analysis} and (2) \textit{multi-objective optimization}.

\subsection{Dataset description}\label{subsec:dataset_description}

\begin{figure}[h]
    \centering
    \includegraphics[width=\linewidth]{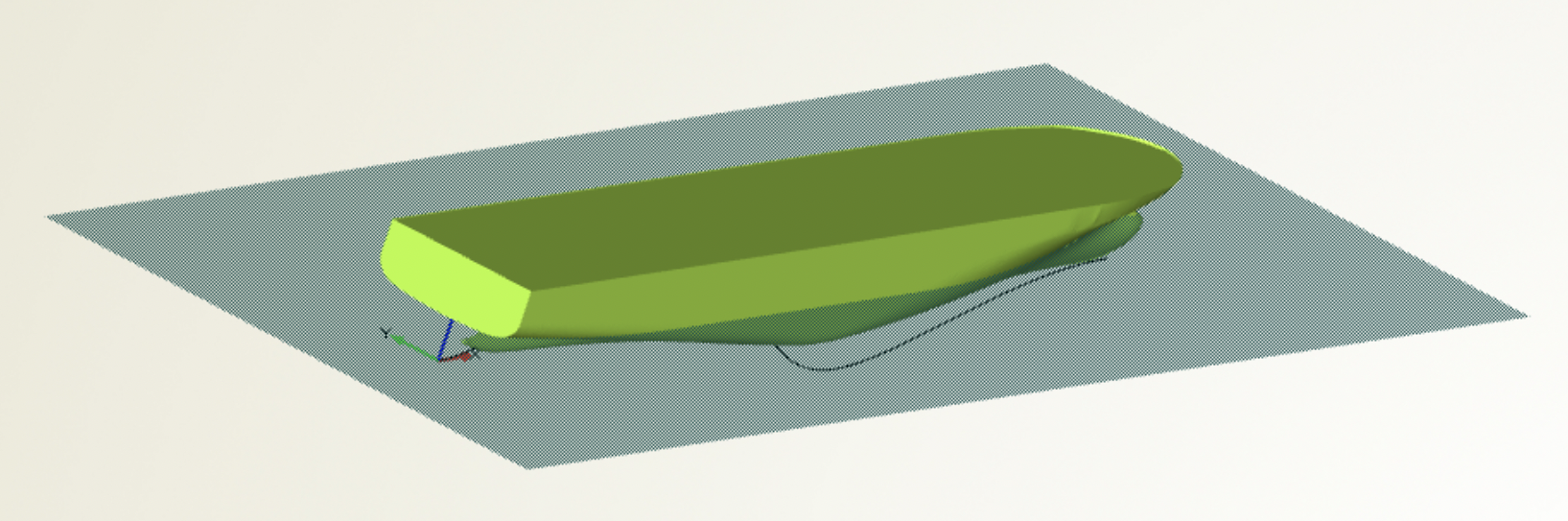}%
    \caption{
    Parametric model of the Duisburg Test Case (DTC) container vessel. Four standard transformations are created, 
    allowing to form complex modifications: scaling in all directions and shifting in the longitudinal direction. A scaling transformation alters the size of geometry whilst a shift transformation changes any point of the geometry by adding a certain displacement depending on the point’s initial position. This displacement is specified as a curve function (highlighted).}\label{fig:parametric-model}
\end{figure}

The dataset contains 140 simulations, carried out using the commercial software \textit{Star-CCM+} \parencite{siemens2021}, including a parametric model in \textit{CAESES} \parencite{caeses2022} designed by Bremen University of Applied Sciences (HSB).
The dataset is characterized by a range of design variations, loading conditions, and velocities. It is split into three sets: (i) the train set that contains 120 simulations used to train the DLP model, (ii) the validation set that contains 10 simulations used to tune the parameters of the DLP model, and (iii) the test set that contains 10 simulations used to assess the predictive performances of the DLP model on new, unseen geometries and conditions. See later on, the reported results in Table \ref{tab:dlp_score}. In this work, we consider the Duisburg Test Case (DTC) container vessel \parencite{elmoctar2012} as the baseline geometry. To meet ship stability requirements, the parameters are only allowed to be changed within a predefined range. 
In Figure \ref{fig:parametric-model}, we depict the baseline geometry.

%

\subsection{DLP performances}\label{subsec:dlp_exp}

The performances of the DLP model are evaluated by comparing the predictions of the test set (not seen by the DLP model during the training) with the corresponding CFD solutions. The first comparison is carried out for the integrated quantities and the second comparison is related to the assessment of the field quantities.


\noindent {\bf Integrated quantities}. We present the results of the DLP model by considering the total forces
as reported in Table \ref{tab:dlp_score}. We recall that these physical quantities have not been seen by the DLP model but are computed as post-treatments by integration on the hull surface. Table \ref{tab:dlp_score} shows that the DLP model achieves high-quality results and can recover the distribution of forces on the hull.

\begin{table}[h]
    \centering
    \caption{Performances of DLP model on the test set of DTC container vessel. $. \pm .$ stands for mean and standard deviation.}\label{tab:dlp_score}
    \begin{tabular}{ p{4cm}||p{4.4cm}}
        \bf Physical quantities & \bf Relative L1 norm in \%  \\
        \hline
        Total Force X &  3.849 $\pm$ 2.179 \\ 
        Total Force Y &  3.414 $\pm$ 1.776 \\ 
        Total Force Z &  2.240 $\pm$ 1.126 \\ 
    \end{tabular}
\end{table}
 

\newpage

\begin{figure}[h]
    \centering\offinterlineskip
    \begin{subfigure}[t]{\linewidth}
        \includegraphics[width=0.5\linewidth]{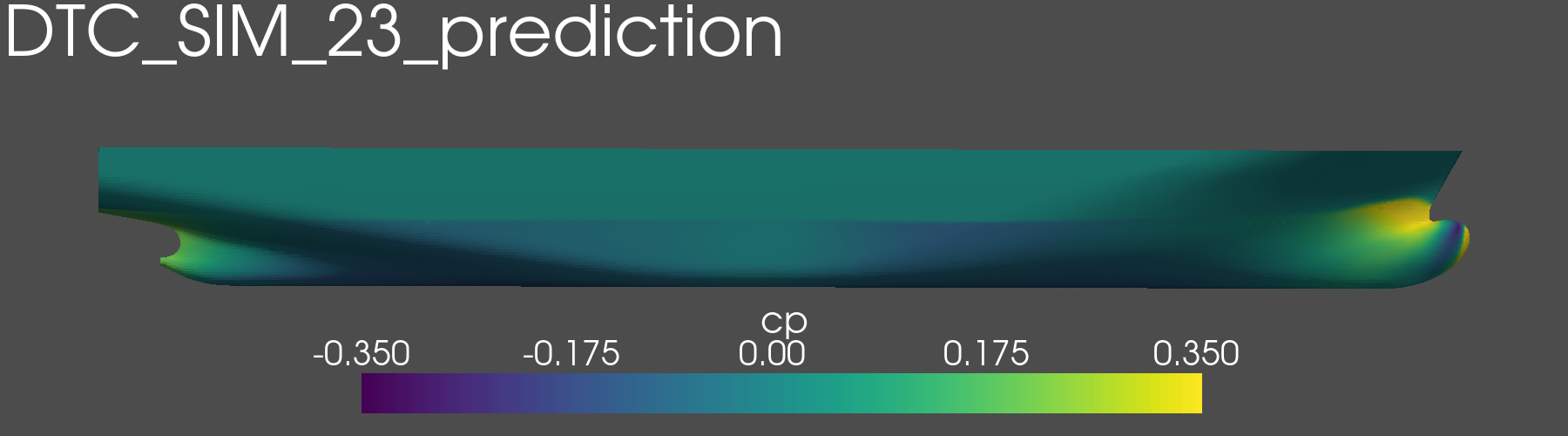}%
        \includegraphics[width=0.5\linewidth]{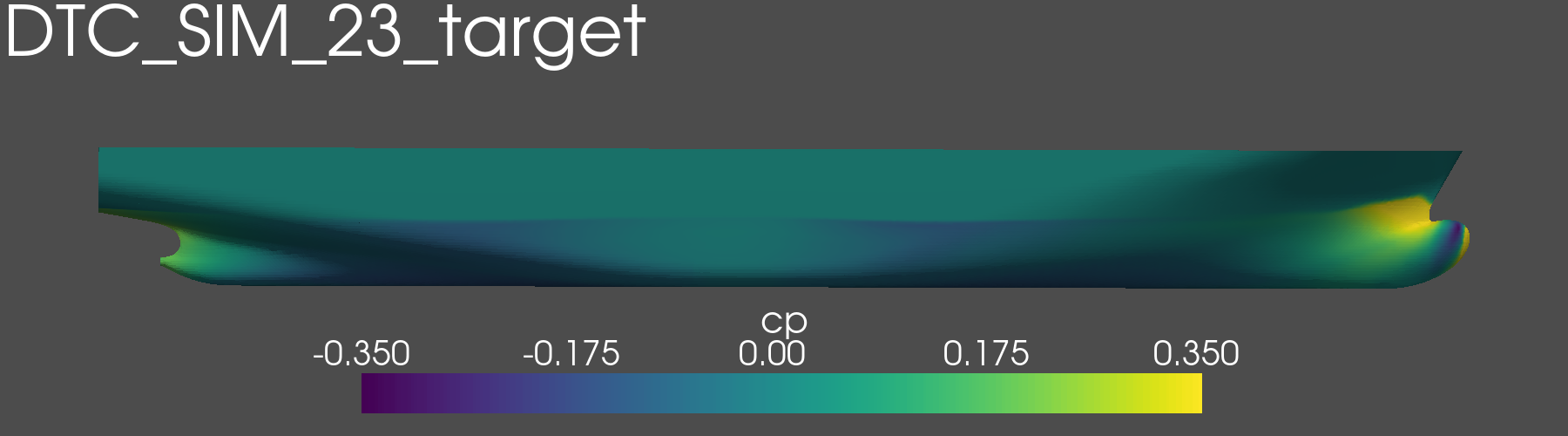}%
        \caption{The distribution displayed is the non-dimensional hydrodynamic pressure coefficient, which is defined as $c_p = 2 (\overline{p} + \rho g z - p_{atm}) / (\rho V_{\infty}^2)$, where the reference pressure, $p_{atm}=1$ $atm$ was used.}
    \end{subfigure}\\
    \begin{subfigure}[t]{\linewidth}
        \includegraphics[width=0.5\linewidth]{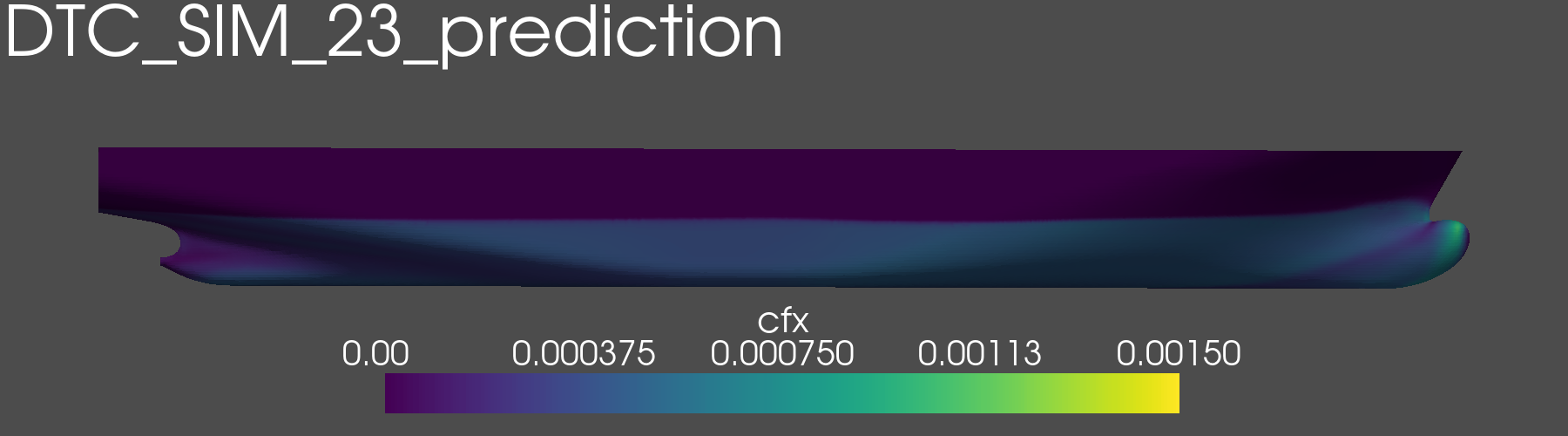}%
        \includegraphics[width=0.5\linewidth]{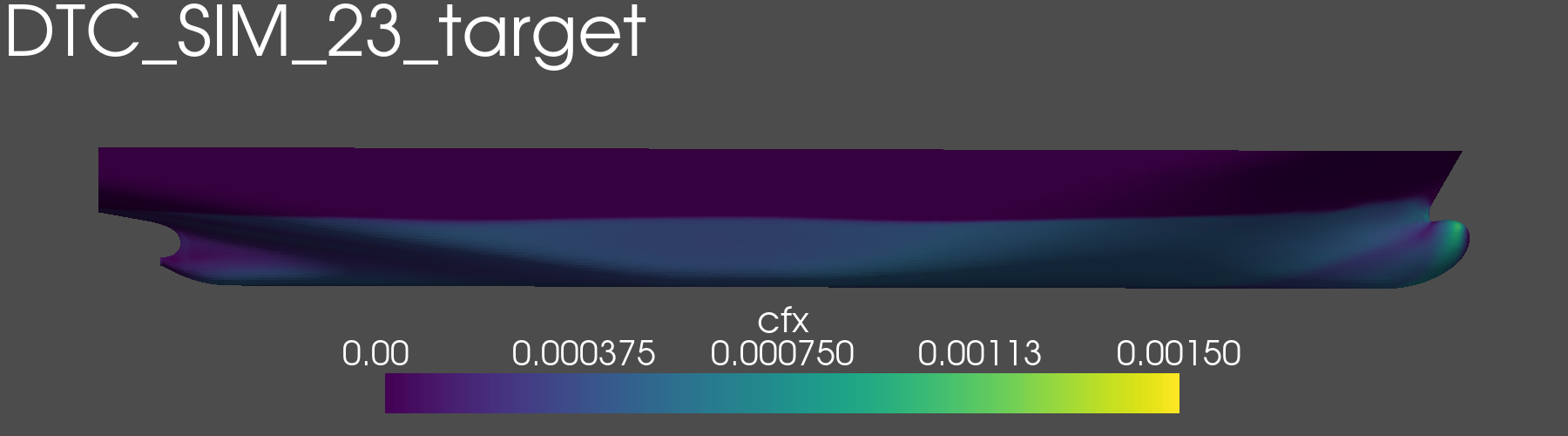}%
        \caption{The distribution displayed is the non-dimensional skin friction coefficient, which is defined as $c_{fx} = 2 \overline{\kappa}_{x} / (\rho S V_{\infty}^2)$, where $S$ is the wetted surface area.}
    \end{subfigure}%
    \caption{Qualitative comparison of the surface distribution of different variables. Left: DLP prediction. Right: CFD result.}\label{fig:results-surface}
\end{figure}

\begin{figure}[H]
    \centering\offinterlineskip
    \includegraphics[width=0.5\linewidth]{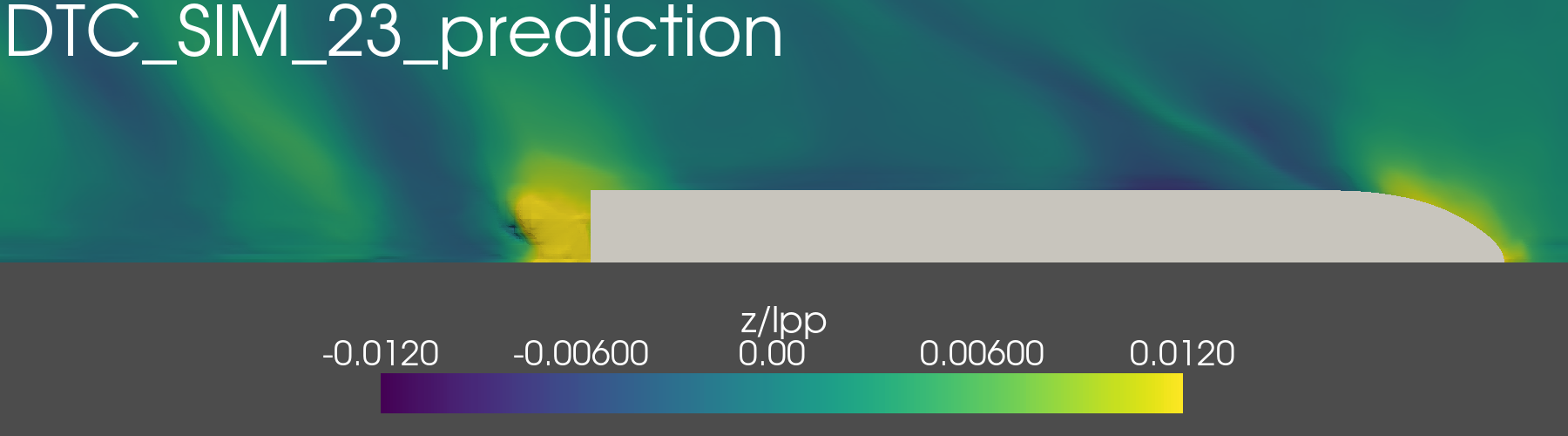}%
    \includegraphics[width=0.5\linewidth]{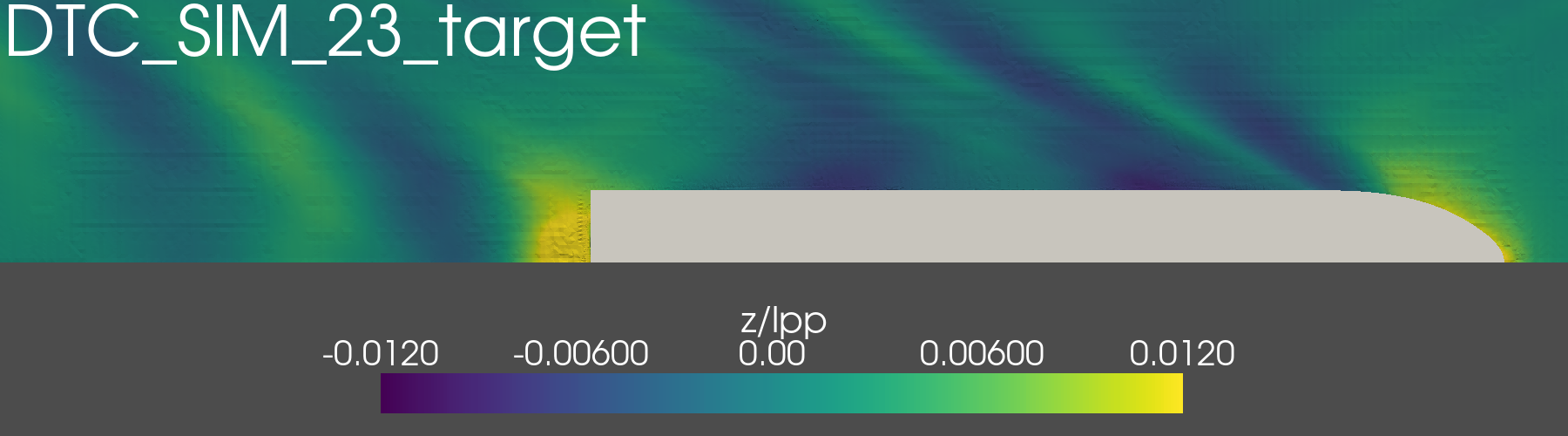}%
    \caption{Iso-surface of water volume fraction = 0.5 colored by the non-dimensional free surface elevation. $L_{PP}$ is the length between perpendiculars. Left: DLP prediction. Right: CFD result.}\label{fig:results-volume}
\end{figure}

\noindent {\bf Field quantities}. Figure \ref{fig:results-surface} illustrates one of the configurations in the test set. We observe a good qualitative agreement between the DLP model and CFD. The DLP model also makes predictions on the full 3D volume where some comparisons can be made within different slices, as depicted in Figure \ref{fig:results-volume}. 
It exhibits the wave elevation at the free surface.
It also shows a good agreement of the wave pattern between the DLP model and CFD. As a consequence, we argue that DLP in combination with a CAD software in an optimization loop could be a good candidate to make shape deformation and optimization. For that reason, in section \ref{subsec:exp_shape_optim}, we propose two scenarios to illustrate our claims (discussed in \ref{subsec:shape_optim}) regarding the potential of DLP to carry optimization studies.

\subsection{Optimization results}\label{subsec:exp_shape_optim}


\noindent {\bf Sensitivity analysis}. 
During the exploration phase, a number of shape variants and the resistance at one freestream velocity is calculated and stored. For these evaluations, a DLPO framework is created, linking the DLP model, a CAD software and an optimizer. 
A Python script is used to communicate commands to the DLP model, in order to: 
(i) incorporate variable parameters, such as the desired conditions and relative file locations, which allows the DLP model to locate the geometry to import, 
(ii) run a DLP prediction and (iii) save the result values.
To perform the results analysis, the result values are used as evaluation parameters to drive the optimization process. Similarly, it is also possible to extract and analyze the wave profiles, near-field wave pattern, and velocities, as well as pressures anywhere in the volume of simulation for each iteration. By changing the value of each parameter, a detailed prediction of the 3D flow around the vessel can be obtained for any combination of these parameters.
Each iteration takes only 20 seconds - including geometry creation, DLP prediction and 
determination of new parameters - which is equivalent in quality with a CFD simulation performed over 32 CPUs during 6 hours. This is more than a thousand times faster, leading to a reduction of computational time and effort 
while keeping accurate and high-quality results. 

\begin{figure}[h]
    \includegraphics[width=\linewidth]{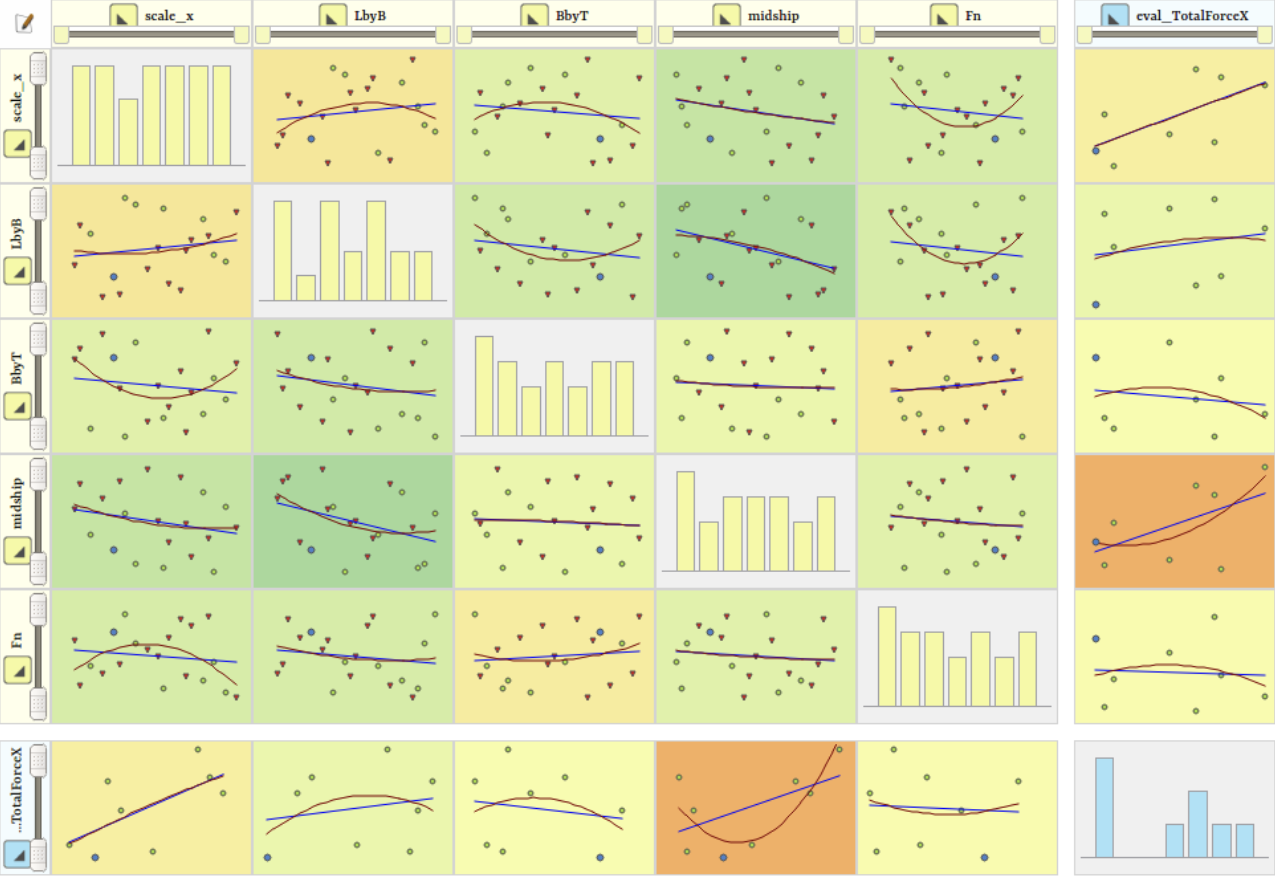}%
    \caption{Scatter plots and trends. In this study, the variables are: the scale factor in x-direction \textit{scale\_x}, length over beam ratio \textit{LbyB}, beam over draught ratio \textit{BbyT}, miship position \textit{midship} and freestream velocity $V_{\infty}$. These variables are changed according to a pre-determined sampling plan. Thus, changing the total resistance $eval\_{TotalForceX}$ estimated with the DLP model. The color background indicates the correlation; red means positive correlation and green means negative correlation. A positive correlation means that the variables move in the same direction, while a negative correlation means that the variables move in opposite directions. The gradient background indicates the sensitivity; dark means more sensitive and light means less sensitive.}\label{fig:boat5}
\end{figure}

Since both input and output are available, it is possible to draw a scatter plot, showing the correlation between the total resistance and the input parameters. In Figure \ref{fig:boat5}, the results show the linear regression (blue line), and the polynomial regression (red line) of the DLP model output versus the shape parameters. 
The study also shows the correlation between all combinations of parameter and the associated resistance. The color background indicates the correlation; red means positive correlation and green means negative correlation. A positive correlation means that the variables move in the same direction, while a negative correlation means that the variables move in opposite directions. The gradient background indicates the sensitivity; dark means more sensitive and light means less sensitive. 
This information 
can be used to change the input such that a hull shape with a better performance can be obtained. 
At this stage, the blue dot already indicates the position of the shape parameters that gives the minimum resistance.

\newpage

\begin{figure}[h]
    \includegraphics[width=\linewidth]{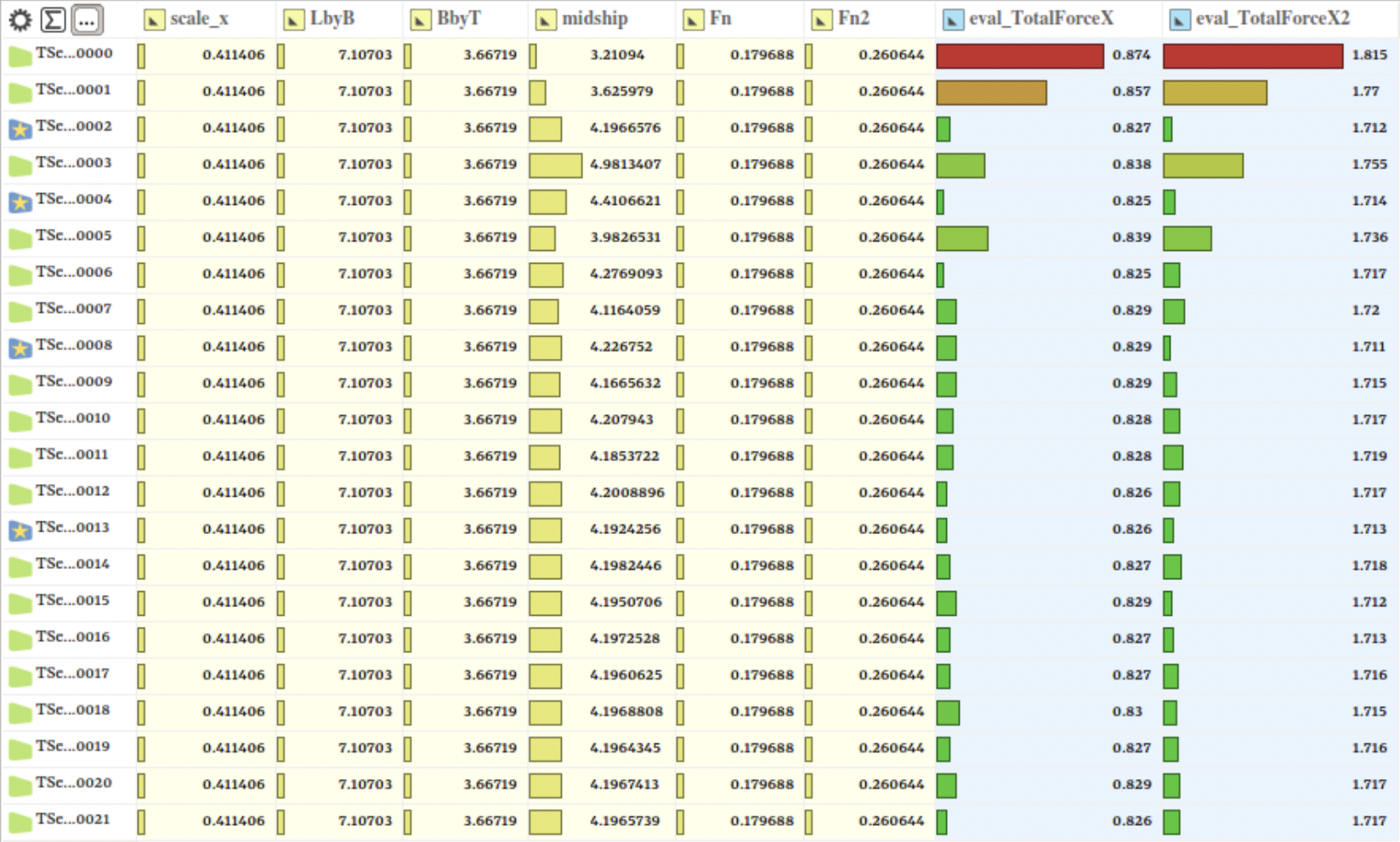}%
    \caption{Results of the multi-objective optimization. In this case, the midship position \textit{midship} is changed, and the optimization progresses towards
    the minimum resistance $eval\_{TotalForceX}$ and $eval\_{TotalForceX2}$ at two speeds
    $Fn$ and $Fn2$. The other variables are frozen.}\label{fig:multi-objective-optimization}
\end{figure}

\begin{figure}[h]
    \centering
    \includegraphics[width=0.8\linewidth]{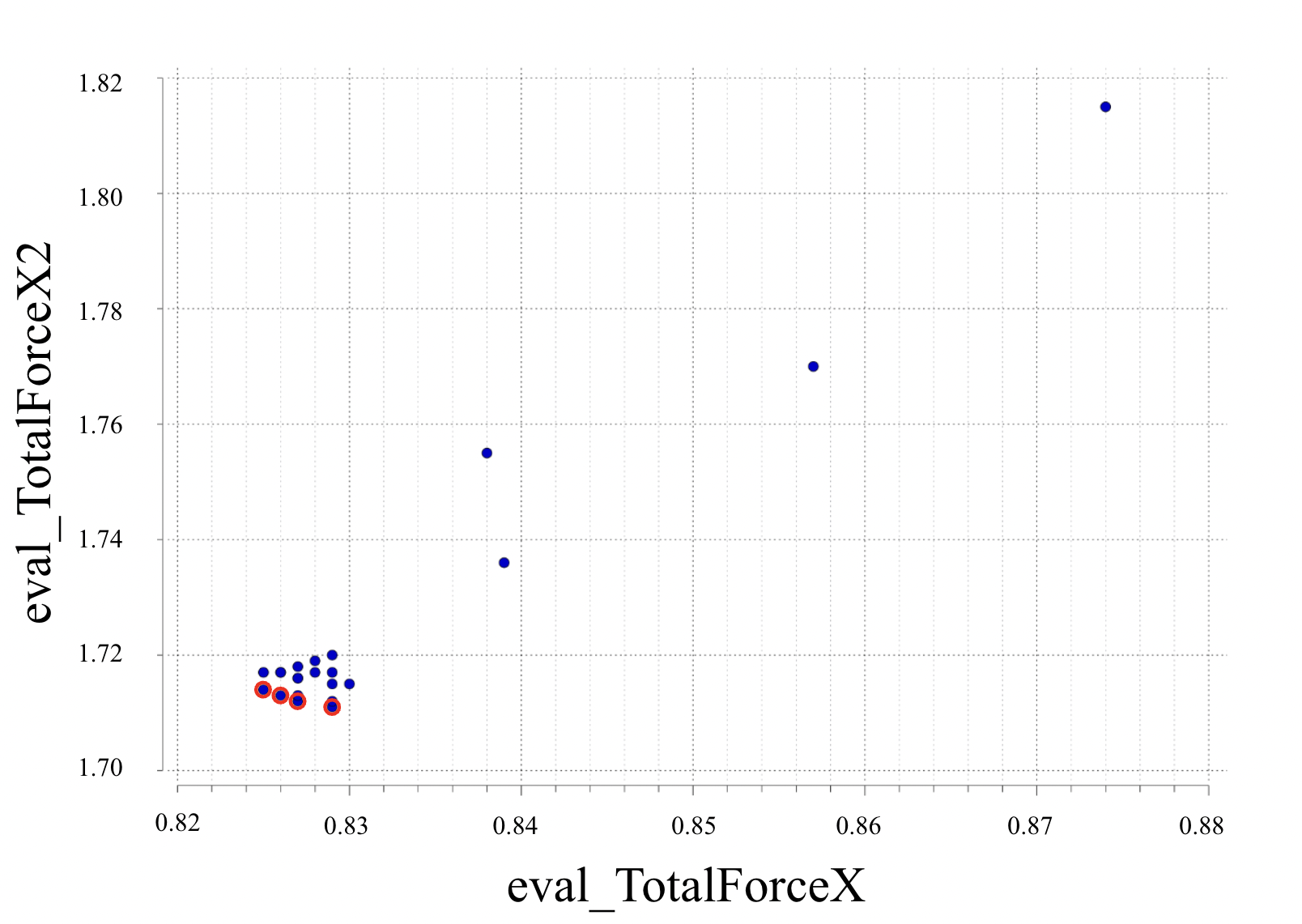}%
    \caption{Pareto front (in red), showing the trade-off between conflicting design objectives. It illustrates that it is not possible to have the lowest resistance for both speeds.}\label{fig:pareto-front}
\end{figure}

\noindent {\bf Multi-objective optimization}. Conversely,
it is also possible to drive the design exploration by systematically changing the shape parameters in order to iteratively convergence towards
an optimum. 
At this stage, 
it is possible to search for two or more objectives at the same time, for example the total resistance at two specific speeds. The results of the multi-objective optimization are shown in Figure \ref{fig:multi-objective-optimization}. Based on the T-Search algorithm (see again section \ref{subsec:shape_optim}), the total resistance at two speeds $Fn=0.18$ and $Fn2=0.26$ has been calculated with multiple values of the midship position. Optimal design points are starred (creating a \textit{Pareto front}, that is, a set of points which are all better w.r.t both objectives). Based on this Pareto front and weight factors for the different speeds, the optimal design point can be found for a particular operational profile. The results are also shown graphically in Figure \ref{fig:pareto-front}, with the horizontal axis being the resistance at the lowest speed and the vertical axis being the resistance at high speed. The points to the upper right quadrant of the plot show design points that are less favorable than the points lying on the Pareto front (in red) w.r.t both objectives. As a final step, to 
validate the optimal design variants, a new CFD simulation can be done, 
which can be used to retrain the model.

\section{CONCLUSIONS}\label{sec:conclusions}


In this work, we proposed an optimization framework that we called \textit{Deep Learning Physics Optimization (DLPO)} to study the optimization of new container vessels more efficiently w.r.t a set of pre-defined constraints and parameters. It is the result of linking Extrality's \textit{Deep Learning  Physics (DLP)} model with a CAD engine and an optimizer. 
The proposed DLP model is trained on full 3D volume data coming from RANS CFD and it shows high-quality results at all scales, ranging from total forces to 3D flow fields. In particular, it achieves a low error prediction of $3.84\% \pm 2.179\%$ on the total resistance.  Therefore, it has opened new perspectives to provide high-quality 3D flow predictions in real-time, including RANS-like detailed flow information.

From a design perspective, we set up two optimization scenarios: (1) \textit{sensitivity analysis} and (2) \textit{multi-objective optimization}. Sensitivity analysis is used to identify the impact that geometry changes have on the hydrodynamic efficiency of the vessel. Multi-objective optimization is used to obtain a Pareto front, where the optimal solutions lie in terms of low resistance for the different speeds. Other results could be considered during the optimization phase. For instance, the analysis of the wave profiles and near-field wave patterns. Each iteration takes only 20 seconds, including geometry creation, 
DLP prediction and determination of new parameters. These two optimization scenarios confirm the capabilities of DLPO to help engineers to optimize design cycles and iterate efficiently as much as needed on their designs.

\newpage

\noindent {\bf Open directions}. 
Our proposed DLPO framework is not only restricted to this design study. It could also support geometries for which there exist no parametric representation in this design space thanks to its geometry-centric approach. In this case, the DLP model could be refined by filling the design space with variations of a new baseline geometry so that the results obtained will be as accurate as possible. 
The same DLP architecture could also be used to set up a shape optimization study for an entirely different ship type. Finally, since most CAD tools provide the ability to use script files to manipulate the geometry, the scripts developed in the present work could be used to, relatively quickly, couple another CAD modeler to the DLP model in an optimization loop, making DLPO framework a fast and flexible optimization tool.

\section*{REFERENCES}


\printbibliography[heading=none]

@article{scholcz2015,
author = {Scholcz, Thomas P. and Gornicz, Tomasz and Veldhuis, Christian},
doi = {10.13140/RG.2.1.4173.5124},
number = {June},
pages = {1064-1077},
title = {Multi-objective hull-form optimization using Kriging on noisy computer experiments},
year = {2015}
}

@article{scholcz2017,
author = {Scholcz, Thomas P. and Veldhuis, Christian H.J.},
journal = {7th International Conference on Computational Methods in Marine Engineering, MARINE 2017},
number = {May},
pages = {231-242},
title = {{Multi-objective surrogate based hull-form optimization using high-fidelity rans computations}},
volume = {2017-May},
year = {2017}
}

@article{queipo2005,
author = {Queipo, Nestor V. and Haftka, Raphael T. and Shyy, Wei and Goel, Tushar and Vaidyanathan, Rajkumar and {Kevin Tucker}, P.},
doi = {https://doi.org/10.1016/j.paerosci.2005.02.001},
issn = {0376-0421},
journal = {Progress in Aerospace Sciences},
number = {1},
pages = {1-28},
title = {Surrogate-based analysis and optimization},
volume = {41},
year = {2005}
}

@article{vanstraten2019,
author = {Straten, Oscar F.A. Van and Celik, Egemen and de Baar, Jouke H.S. and Ascic, Blanka and de Jong, Jochem S.},
isbn = {9788494919435},
journal = {8th International Conference on Computational Methods in Marine Engineering, MARINE 2019},
number = {4},
pages = {321-333},
title = {Improved hull design with potential-flow-based parametric computer experiments},
volume = {4},
year = {2019}
}

@misc{yu2018,
title={Hull Form Optimization with Principal Component Analysis and Deep Neural Network}, 
author={Dongchi Yu and Lu Wang},
year={2018},
eprint={1810.11701},
archivePrefix={arXiv},
primaryClass={stat.ML}
}

@article{hornik1989,
title = {Multilayer feedforward networks are universal approximators},
journal = {Neural Networks},
volume = {2},
number = {5},
pages = {359-366},
year = {1989},
issn = {0893-6080},
doi = {https://doi.org/10.1016/0893-6080(89)90020-8},
author = {Kurt Hornik and Maxwell Stinchcombe and Halbert White}
}

@article{elmoctar2012,
author = {El Moctar, Ould and Shigunov, Vladimir and Zorn, Tobias},
doi = {10.1179/str.2012.59.3.004},
issn = {09377255},
journal = {Ship Technology Research},
number = {3},
pages = {50-64},
title = {Duisburg test case: Post-panamax container ship for benchmarking},
volume = {59},
year = {2012}
}

@article{vandenboogaard2022,
author = {Boogaard, Maurits Van Den and Alessi, Giacomo and Mallol, Benoit and Wunsch, Dirk and Clero, Nathan},
journal = {21th Conference on Computer Applications and Information Technology in the Maritime Industries, COMPIT 2022},
number = {May},
pages = {7-15},
title = {Accelerating marine propeller development in early design stages using machine learning},
year = {2022}
}

@software{siemens2021,
author = {Siemens Digital Industries, Software},
title = {Simcenter \uppercase{STAR-CCM+}},
url = {https://www.plm.automation.siemens.com/global/en/products/simcenter/STAR-CCM.html},
version = {2021.1},
year = {2021}
}

@software{caeses2022,
author = {Friendship, Systems},
title = {\uppercase{CAESES}},
url = {https://www.friendship-systems.com/products/caeses/},
version = {4.4},
year = {2018}
}

@article{sha1994,
author = {Sha, O.P. and Ray, T. and Gokarn, R.P.},
journal = {8th International Conference on Computer Applications in Shipbuilding, ICCAS 94},
pages = {10-15},
title = {An artificial neural network model for preliminary ship design},
volume = {2},
year = {1994}
}

@article{pfaff2020,
archivePrefix = {arXiv},
arxivId = {2010.03409},
author = {Pfaff, Tobias and Fortunato, Meire and Sanchez-Gonzalez, Alvaro and Battaglia, Peter W.},
eprint = {2010.03409},
pages = {1-18},
title = {Learning Mesh-Based Simulation with Graph Networks},
url = {http://arxiv.org/abs/2010.03409},
year = {2020}
}

@article{bronstein2017,
author = {Bronstein, Michael M. and Bruna, Joan and LeCun, Yann and Szlam, Arthur and Vandergheynst, Pierre},
journal = {IEEE Signal Processing Magazine}, 
title = {Geometric Deep Learning: Going beyond Euclidean data}, 
year = {2017},
volume = {34},
number = {4},
pages = {18-42},
doi = {10.1109/MSP.2017.2693418}
}

@book{forrester2018,
author = {Forrester, Alexander I. J. and Sobester, Andras and Keane, Andy J.},
isbn = {978-0-470-06068-1},
pages = {I-XVIII, 1-210},
publisher = {Wiley},
title = {Engineering Design via Surrogate Modelling - A Practical Guide},
year = {2008}
}

@book{letcher2009,
title = {The Geometry of Ships},
author = {Letcher, J.S. and Paulling, J.R. and Society of Naval Architects and Marine Engineers (U.S.)},
isbn = {9780939773671},
lccn = {2012382391},
series = {The principles of naval architecture series},
url = {https://books.google.de/books?id=gEHUMgEACAAJ},
year = {2009},
publisher = {Society of Naval Architects and Marine Engineers}
}

@inproceedings{raven2017,
author = {Raven, H.C. and Scholcz, Thomas},
title = {Wave resistance minimisation in practical ship design},
booktitle = {7th International Conference on Computational Methods in Marine Engineering, MARINE 2017},
pages = {},
volume = {2017-May},
year = {2017}
}

@misc{kingma2014,
author = "Kingma, Diederik P. and Ba, Jimmy",
title = "{Adam: A Method for Stochastic Optimization}",
eprint = "1412.6980",
archivePrefix = "arXiv",
primaryClass = "cs.LG",
year = "{2014}"
}

@article{sobol1988,
title = {On quasirandom sequences for numerical computations},
journal = {USSR Computational Mathematics and Mathematical Physics},
volume = {28},
number = {3},
pages = {88-92},
year = {1988},
issn = {0041-5553},
doi = {https://doi.org/10.1016/0041-5553(88)90181-4},
author = {Yu.L. Levitan and N.I. Markovich and S.G. Rozin and I.M. Sobol}
}

@inproceedings{hilleary1966,
title={The tangent search method of constrained minimization},
author={Rober R. Hilleary},
year={1966}
}

@online{britannica2018,
author = {Conn, J.F.C. and Pounder, Cuthbert Coulson},
title = {Encyclopedia Britannica: ship construction},
url = {https://www.britannica.com/technology/ship-construction},
year = {2018}
}

@inproceedings{krizhevsky2012,
 author = {Krizhevsky, Alex and Sutskever, Ilya and Hinton, Geoffrey E},
 booktitle = {Advances in Neural Information Processing Systems},
 editor = {F. Pereira and C.J. Burges and L. Bottou and K.Q. Weinberger},
 pages = {},
 publisher = {Curran Associates, Inc.},
 title = {ImageNet Classification with Deep Convolutional Neural Networks},
 url = {https://proceedings.neurips.cc/paper_files/paper/2012/file/c399862d3b9d6b76c8436e924a68c45b-Paper.pdf},
 volume = {25},
 year = {2012}
}







\end{document}